\pdfoutput=1

\documentclass[11pt]{article}

\usepackage[preprint]{acl}

\usepackage{times}
\usepackage{latexsym}
\usepackage{tabularx}
\usepackage[T1]{fontenc}

\usepackage[utf8]{inputenc}

\usepackage{microtype}

\usepackage{inconsolata}

\usepackage{graphicx}
\usepackage[ruled]{algorithm2e} 
\usepackage{amsmath}
\usepackage{amssymb}
\usepackage{enumitem}
\usepackage{pifont}
\usepackage{tikz,pgfplots}
\usepackage{adjustbox}
\usepackage{booktabs}
\usepackage{colortbl}
\usepackage{xcolor}
\usepackage{natbib}

\newcommand{\ours}{\textsc{AlphaLLM-CPL}\xspace}

\newcommand{\eg}[0]{\emph{e.g., }}

\newcommand{\RN}[1]{%
	\textup{\lowercase\expandafter{\it \romannumeral#1}}%
}

\usepackage{amsmath,amsfonts,bm}

\def\eqref#1{equation~\ref{#1}}

\def\1{\bm{1}}

\def\vx{{\bm{x}}}
\def\vy{{\bm{y}}}

\DeclareMathAlphabet{\mathsfit}{\encodingdefault}{\sfdefault}{m}{sl}
\SetMathAlphabet{\mathsfit}{bold}{\encodingdefault}{\sfdefault}{bx}{n}

\def\gA{{\mathcal{A}}}

\def\gD{{\mathcal{D}}}

\def\gS{{\mathcal{S}}}

\title{Towards Self-Improvement of LLMs via MCTS: \\ Leveraging Stepwise Knowledge with Curriculum Preference Learning}

\author{Xiyao Wang$^{\ 1 *}$\quad Linfeng Song$^{\ 2}$ \quad Ye Tian$^{\ 2}$ \quad Dian Yu$^{\ 2}$ \quad Baolin Peng$^{\ 2}$\quad \\ 
\textbf{Haitao Mi$^{\ 2}$} \quad \textbf{Furong Huang$^{\ 1}$} \quad \textbf{Dong Yu$^{\ 2}$} \\
         $^{1}$ University of Maryland, College Park \\ $^{2}$ Tencent AI Lab, Bellevue, WA \\ \texttt{xywang@umd.edu} }
\begin{document}
\maketitle
\begingroup\renewcommand\thefootnote{*}
\footnotetext{Work done during an internship at Tencent AI Lab}
\begin{abstract}
Monte Carlo Tree Search (MCTS) has recently emerged as a powerful technique for enhancing the reasoning capabilities of LLMs. Techniques such as SFT or DPO have enabled LLMs to distill high-quality behaviors from MCTS, improving their reasoning performance. However, existing distillation methods underutilize the rich trajectory information generated by MCTS, limiting the potential for improvements in LLM reasoning. 
In this paper, we propose \ours, a novel pairwise training framework that enables LLMs to self-improve through MCTS behavior distillation. \ours efficiently leverages MCTS trajectories via two key innovations: (1) \ours constructs stepwise trajectory pairs from child nodes sharing the same parent in the search tree, providing step-level information for more effective MCTS behavior distillation. (2) \ours introduces curriculum preference learning,  dynamically adjusting the training sequence of trajectory pairs in each offline training epoch to prioritize critical learning steps and mitigate overfitting. Experimental results on mathematical reasoning tasks demonstrate that \ours significantly outperforms previous MCTS behavior distillation methods, substantially boosting the reasoning capabilities of LLMs.
\end{abstract}

\section{Introduction}
\label{sec: intro}

Recent work has shown that large language models (LLMs), trained on trillions of tokens and comprising billions of parameters, exhibit extraordinary capabilities
across a wide range of natural language processing tasks \citep{achiam2023gpt,touvron2023llama,team2023gemini}.
Nevertheless, they still face challenges in tasks requiring rigorous reasoning and planning, such as math-problem solving \citep{huang2023large,valmeekam2024planbench}.
Although Chain-of-Thought (CoT) prompting \citep{wei2022chain,kojima2022large}, which encourages LLMs to generate full reasoning steps before arriving at an answer, has shown promising results, LLMs continue to struggle with problems requiring extended reasoning steps due to the limitations of auto-regressive decoding. 
Recent work suggests finetuning LLMs using either a substantial amount of high-quality, supervised data \citep{yumetamath,yuan2023scaling,li2024common,mishra2022lila, tong2024dartmathdifficultyawarerejectiontuning}, guidance from a stronger LLM (e.g. GPT4) \citep{gou2023tora,lee2023platypus}, or a combination of both \citep{luo2023wizardmath,yue2023mammoth}.
However, these approaches are constrained by the high cost of extensively querying stronger LLMs and the limited scope and quality of data that humans can provide.

In response to these challenges, the use of Monte Carlo Tree Search (MCTS) within LLMs has garnered increasing attention. 
By leveraging LLMs as the policy for MCTS in the language space, it becomes possible to generate high-quality data without querying stronger models or relying on human annotation. 
The high-value behaviors identified through MCTS are subsequently distilled into the LLM’s policy via supervised fine-tuning (SFT) or direct preference optimization (DPO)~\citep{rafailov2024direct}, enabling self-improvement of the LLM. 
This methodology has demonstrated considerable potential in recent studies, significantly enhancing the reasoning capabilities of LLMs and proving effective in complex reasoning tasks.

However, existing MCTS distillation methods have notable limitations. 
In SFT-based distillation, only the best trajectory found by MCTS is typically used as training data, while other potentially useful data generated during the search is discarded~\citep{tian2024toward,zhang2024rest}. 
Although DPO-based distillation leverages more trajectories as trajectory pairs, it still underutilizes the supervisory information provided by MCTS to optimize the training process~\citep{xie2024montecarlotreesearch}. 
As a result, LLMs fail to fully exploit the trajectories generated by MCTS, requiring multiple rounds of online tree search to continually acquire new samples for improving model performance.
\begin{figure*}[!htbp]
    \centering
    \includegraphics[width=\linewidth]{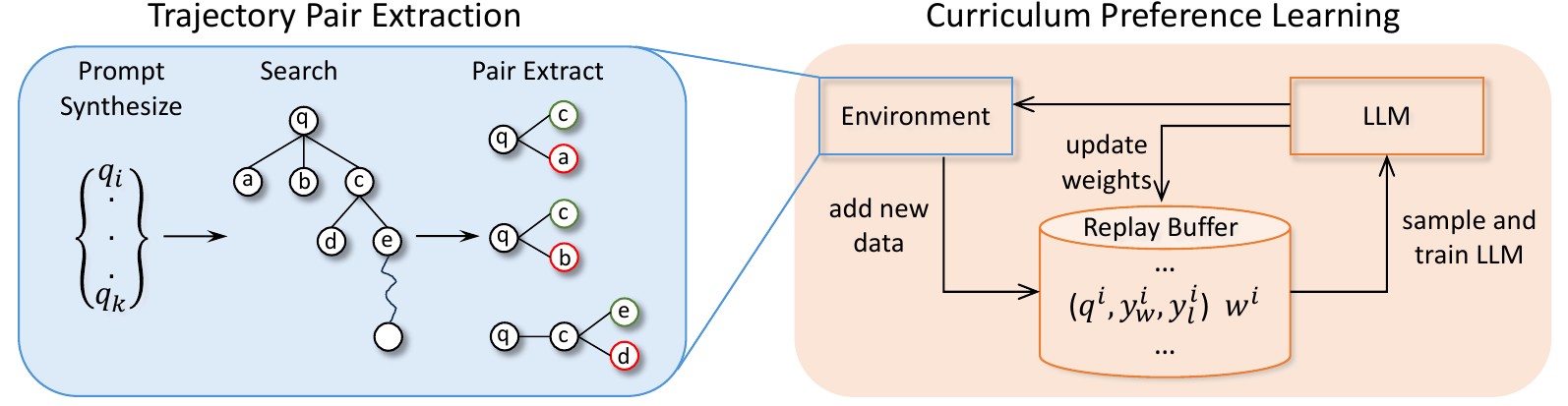}
    \caption{\ours loop: We first performs MCTS over each input prompt $q$ before extracting stepwise trajectory pairs, and then construct all extracted trajectory pairs into an offline replay buffer. We propose curriculum preference learning to continuously update the weight of each trajectory pair in the replay buffer during each offline training epoch and update the LLM.}
    \label{fig:flow}
\end{figure*}

In this paper, we propose \ours, a pairwise offline training framework designed to enable LLMs to self-improve through MCTS behavior distillation. 
As illustrated in Figure \ref{fig:flow}, \ours consists of two key components:
\textbf{(a) Trajectory Pair Extraction.} 
After MCTS generates multiple trajectories for a given prompt, we form complete trajectory pairs based on their respective trajectory values. 
Additionally, we extract stepwise trajectory pairs by selecting child nodes from the same root node in the search tree.
These stepwise trajectory pairs provide step-level information that enhances MCTS behavior distillation.
Our experiments show that incorporating stepwise trajectory pairs not only stabilizes the preference learning process but also significantly improves LLM performance compared to using only complete trajectory pairs. 
\textbf{(b) Curriculum Preference Learning.}
We construct a replay buffer of the extracted trajectory pairs. 
At each offline training epoch, we dynamically adjust the sequence of trajectory pairs in this replay buffer, prioritizing those more critical to the LLM's learning.
The adjustment is based on the preference reward gap provided by MCTS reward model and policy prediction gap provided by LLM policy itself. 
Compared to a random sampling of trajectory pairs for preference learning, curriculum preference learning consistently improves the LLM's reasoning capabilities across multiple offline training epochs.
Together, these two components enable \ours to more efficiently utilize the trajectories generated by MCTS, allowing the LLM policy to comprehensively distill and learn from MCTS behavior.

We conduct experiments on two mathematical reasoning tasks, GSM8K and MATH. For the GSM8K benchmark, we select LLaMA2-7B and Mistral-7B as our base models. For the more challenging MATH benchmark, we use the stronger LLaMA3-8B-Instruct as the base model. 
After applying \ours, the performance of LLaMA2-7B and Mistral-7B on GSM8K improve from 14.6 and 38.5 to 36.5 and 57.3, representing gains of \textbf{150\%} and \textbf{48.8\%}, respectively. 
On the more difficult MATH benchmark, \ours increases the performance of LLaMA3-8B-Instruct from 28.2 to 33.1, achieving a \textbf{17.4\%} improvement. 
Additionally, \ours demonstrates a clear advantage over other MCTS behavior distillation methods~\citep{tian2024toward, xie2024montecarlotreesearch} in terms of performance.

\section{Preliminaries}
\label{sec: prelim}

\subsection{Problem Formulation}

Formally, given a query $q=[q^1,\dots,q^n]$, which is typically referred to as prompt, a LLM denoted as policy $\pi_\theta$ is adopted to generate the response $y=[y^1,\dots,y^m]$.
The policy $\pi_\theta$ operates in an autoregressive manner, where each token is generated sequentially, relying solely on the context provided by the previously generated tokens.
The policy, therefore, constitutes a Markov process in which the conditional probability distribution $\pi_\theta(y|q)$ can be decomposed and expressed with the chain rule:
\begin{equation}
    \pi_\theta(y|q) = \prod_{i=1}^{m} \pi_\theta(y^i|q, y^{<i}; \theta)
\end{equation}
This can be formulated as a Markov Decision Process (MDP) problem, with elements $(\gS, \gA, T, R, \gamma)$. In this structure, the state $s^t \in \gS$ represents the current context information in the trajectory, which is the concatenation of $q$ and $y^{<t}$. The action $y^t \in \gA$ corresponds to a single token selected from the vocabulary, resulting in a transition to a new state by appending $s^t$ and $y^t$, and receiving a reward $r_t = R(s^t, y^t)$.
This MDP framework sets the stage for applying Reinforcement Learning (RL) methods to optimize
the policy $\pi_\theta$ aiming to maximize the expected cumulative reward.
Based on these setups, we describe the self-improving problem.
Given an LLM $\pi_\theta$ and an initial dataset $D_0$, which consists of limited expert-generated prompt-response ($q$, $y$) pairs, the goal of self-improving is
to refine $\pi_\theta$ to maximize the reward.
The refinement process includes learning from
synthesized prompts and corresponding responses, where the responses are obtained using Monte Carlo Tree Search.

\subsection{Monte Carlo Tree Search (MCTS)}

MCTS is a sampling-based search algorithm for policy optimization in decision-making problems. It would iteratively build a search tree, by repeating four phases: selection, expansion, evaluation, and back-propagation.
In the selection phase, it would recursively select the children from the root node by Upper Confidence Bound (UCB) bandit \cite{auer2002finite}, which is
\begin{equation} \label{eq:ucb}
    UCB(i)=w_i + C \times \sqrt{2 \times \ln{\frac{N_i}{n_i}}}
\end{equation}
where $n_i$ and $N_i$ are the visit counts for the node $i$ and its parent respectively, $C$ represents a hyperparameter balancing exploration and exploitation, and the $w_i$ is the average value of all descendant
nodes of $i$.
Following selection, the tree undergoes expansion according to the defined policy in the expansion phase.
Then in the evaluation phase, the value of the newly expanded node is estimated, by sampling or model-based methods.
Finally, in the back-propagation phase, the estimated value is back-propagated to all ancestor nodes of the newly expanded node.

\subsection{Direct Performance Optimization (DPO)}

As an off-policy preference learning algorithm, DPO directly takes pair-wise preference data to tune the policy model $\pi_\theta$ (instead of training a reward model using such data) with an optimization objective equivalent to bandit PPO \cite{schulman2017proximal}.
Particularly, given an input prompt $q$, and a preference data pair ($y_w$, $y_l$), the optimization objective is formulated as:
\begin{multline} \label{eq:dpo}
    \mathcal{L}_{DPO} = -\mathbb{E}_{(q, y_w, y_l) \sim D} \log \sigma \big( \\ \beta \log\frac{\pi_\theta(y_w|q)}{\pi_{ref}(y_w|q)}
    - \beta \log\frac{\pi_\theta(y_l|q)}{\pi_{ref}(y_l|q)} \big)
\end{multline}
where $D$ is the pair-wise preference dataset.
DPO enlarges the gap between the probability of the preferred output $\pi_\theta(y_w|q)$ and that of the loss one $\pi_\theta(y_l|q)$ compared to the reference policy $\pi_{ref}$.
In this way, the policy can learn to better distinguish good responses from bad ones.

\section{AlphaLLM-CPL}\label{sec: alphacpl}

In this section, we will introduce our proposed method AlphaLLM-CPL. AlphaLLM-CPL builds upon the previous self-improvement work, AlphaLLM~\citep{tian2024toward}, and consists of two key components: trajectory pair extraction and LLM self-improvement using curriculum preference learning. 
We will first explain how to generate and construct trajectory pairs for preference learning in Sec~\ref{sec:data}, followed by an introduction to our novel offline preference learning approach, curriculum preference learning, in Sec~\ref{sec:CPL}.
The pseudocode of AlphaLLM-CPL is provided in Algorithm~\ref{algo:self_improving}.

\subsection{Trajectory Pair Extraction}
\label{sec:data}

\paragraph{Executing MCTS on Synthetic Queries}
The first step of data generation is performing Monte-Carlo Tree Search for each input query $q$ that is synthesized from the initial dataset $D_0$.
There have been various strategies, such as Self-instruct~\citep{wang2023self} and Evol-instruct~\citep{luo2023wizardmath}.
In this work, we adopt a method similar to that described in~\citet{yumetamath}.

We follow the MCTS setup in \citet{tian2024toward} except that the search process is only guided by the value network due to computation budget limitation. 
In particular, MCTS performs the following operations iteratively:
\setlength{\leftmargini}{10pt}
\begin{itemize}
    \item \textbf{Selection}~~Starting from the root node, it iteratively selects the child node based on Eq~\ref{eq:ucb}.
    \item \textbf{Expansion}~~Once an expandable leaf node is selected, a new node is generated by starting with the previous state of the parent node as the initial option state. The option is then sampled using the policy $\pi_\theta$, and its completion is determined by a termination function.
    \item \textbf{Backpropagation}~~The average value of the newly generated node and all its ancestors is updated using the scaled reward from the evaluation step. Meanwhile, the visit counts for these nodes also increase by one.
\end{itemize}

\paragraph{Extracting Stepwise Trajectory Pairs from MCTS}
Performing MCTS over query $q$ yields a search tree $T_q$.
The next step is to extract representative data pairs that distinguish good reasoning steps from bad ones.
One intuitive way is to extract all pairs of child nodes from $T_q$, where one yields the correct answer while the other yields the wrong one.
However, this method may prune out many representative pairs since numerous tree nodes may not have the opportunity to roll out to the final answer.
Moreover, an intermediate tree node may yield either a correct or incorrect answer due to the randomness inherent in LLM sampling.

We propose to continue using the value network used by MCTS to select child nodes to form pairs for stepwise comparison.
This is because both recent research \citep{wang2024litesearch} and our findings show that the Q-values are well-calibrated and effectively reflect the quality of deductions that have been made.
Particularly, as shown in Figure \ref{fig:flow}, we empirically set a minimum margin $\tau$ of the Q-value gap, and any pairs of child nodes with a value gap larger than $\tau$ are extracted.

In addition to the stepwise comparison data, we also extract all pairs of complete trajectories based on the same margin $\tau$ for the Q-value gap.
In this case, the two trajectories inside a pair may not share a common prefix.

\subsection{Curriculum Preference Learning}
\label{sec:CPL}

After obtaining trajectory pairs, we consider all trajectory pairs as an offline trajectory pair buffer. 
To more effectively utilize the supervision information provided by MCTS and motivated by previous prioritized experience replay (PER) works in reinforcement learning~\citep{schaul2015prioritized, katharopoulos2018not, wang2023live}, we propose Curriculum Preference Learning (CPL), an offline training method that performs preference learning by continually updating the order of trajectory pairs to select pairs that are more important for policy learning.
We rank the trajectory pairs in the offline buffer based on specific metrics in each offline training epoch, prioritizing training on the more significant pairs. 
Our metric consists of two components: a static part, the preference reward gap, which remains constant across different offline training epochs, and a dynamic part, the policy prediction gap, which evolves as the policy is updated.

\paragraph{Preference reward gap} The preference reward gap is calculated as shown in Eq~\ref{eq: rg}. We leverage the reward model obtained during the pretraining phase to compute the reward for each step in the positive and negative trajectories within a trajectory pair. Then, we calculate the total reward for each trajectory, and the difference between the total rewards of the positive and negative trajectories is used as the ranking weight for each trajectory pair. Since the reward model remains unchanged during the offline training process, the reward gap for each trajectory pair also remains constant with increasing training epochs. The preference reward gap allows the policy to learn the trajectory pairs in the buffer from easy to difficult in each training epoch, improving learning efficiency through a curriculum learning way.

\begin{align}
\label{eq: rg}
r_g(q, y_w, y_l) = \sum_{t=1}^{T_w} R(y_{w, t}, q) - \sum_{t=1}^{T_l} R(y_{l, t}, q)
\end{align}

\paragraph{Policy prediction gap} Policy prediction gap is a policy-related metric that aims to fine-tune the trajectory pair order in each offline training epoch. 
We first calculate the log-likelihood of the positive and negative trajectories in each trajectory pair under the current policy's prediction (Eq~\ref{eq: log}), then compute the difference between them as a metric to rank all preference pairs (Eq~\ref{eq: pg}). 
The trajectory pair with large policy prediction gap means it can provide stronger guidance for policy learning. 
The more distinct the difference between the positive and negative trajectories, the clearer the feedback signal the model can obtain from these samples, allowing it to update parameters more quickly. This enables the model to better learn the distinguishing features between the positive and negative trajectories, thereby improving its performance.
Moreover, as the policy continuously updates, the policy prediction gap for each trajectory pair changes after every offline training epoch. Therefore, this metric dynamically updates the trajectory pair order in the offline trajectory pair buffer, helping the policy converge more efficiently.

\begin{align}
\label{eq: log}
\log P(y \mid q; \theta) = \sum_{t=1}^{T} \log P(y_t \mid q, y_{<t}; \theta) 
\end{align}

\begin{align}
\label{eq: pg}
p_g(q, y_w, y_l) = \log \frac{P(y_w \mid q; \theta)}{P(y_l \mid q; \theta)} 
\end{align}

In the practical implementation, since the scales of the preference reward gap and policy prediction gap are different, we first normalize both the preference reward gap and policy prediction gap to the (0,1) range before adding them together. Thus, the final metric for CPL is the combination of preference reward gap and policy prediction gap with a balance rate $\alpha$: 

\begin{align}
\label{eq: combine}
w_g(q, y_w, y_l) = r_g(q, y_w, y_l) + \alpha * p_g(q, y_w, y_l)
\end{align}

\paragraph{Prompt-level ranking}

To prevent the training process from overly focusing on a fixed prompt—where all trajectory pairs of a specific prompt are prioritized for training, leading to potential overfitting—we employ a prompt-level ranking approach to sort and train the preference pairs. After obtaining the weight coefficients for each trajectory pair, we first internally rank all preference pairs within the same prompt based on these weight coefficients. Then, we select the highest-priority trajectory pair from each prompt, traverse all prompts, and subsequently choose the second-highest priority trajectory pair from each prompt. This process continues iteratively until all response pairs are traversed, and the training follows this sequence in the DPO process.

\begin{algorithm*}[!htb]
\caption{AlphaLLM-CPL}
\textbf{Input} Initial dataset $\gD^0 = \{(\vx_i^0, \vy_i^0) \mid i \in [N]\}$, policy model $\pi_\theta^0$, AlphaLLM-pretrained reward model $R$, number of offline training loop $K$, trajectory value gap margin $\tau$


Generate synthetic prompts $[\hat{\vx}] = \texttt{SYN}(\pi_\theta^0, \gD^0)$

Collect trajectories with search algorithm, \eg MCTS guided by $R$. $[\hat{\vy}] = \texttt{MCTS}(\pi_\theta^0, [\hat{\vx}])$

Construct trajectory pair replay buffer $\hat{\gD} = \{(\hat{\vx}, \hat{\vy}^w, \hat{\vy}^l) \}$ with trajectory value gap margin $\tau$

Policy model $\pi = \pi_\theta^0$

\For{$k \leftarrow 1, \dots, K$}{    
    Sorted preference pair replay buffer $\hat{\gD}_S = []$

    Compute preference reward gap $r_g$ of each trajectory pair in $\hat{\gD}$ using $R$ 

    Compute policy prediction gap $p_g$ of each trajectory pair in $\hat{\gD}$ using $\pi$ 

    \For{ $\hat{x}^i$ in $[\hat{\vx}]$}{
        Select all trajectory pairs in $\hat{\gD}$ that synthetic prompt is $\hat{x}^i$: $\hat{\gD}^i = \{(\hat{x}^i, \hat{\vy}_1^w, \hat{\vy}_1^l), (\hat{x}^i, \hat{\vy}_2^w, \hat{\vy}_2^l), ..., (\hat{x}^i, \hat{\vy}_n^w, \hat{\vy}_n^l) \}$

        Sort trajectory pairs in $\hat{\gD}^i$ according to Eq.(to be defined)
    }
    \While{any $\hat{\gD}^i$ is not empty}{
        \ForAll{$\hat{\gD}^i$}{
            \If{$\hat{\gD}^i$ is not empty}{
                \textbf{Remove} the first trajectory pair from $\hat{\gD}^i$
                
                \textbf{Append} the removed trajectory pair to $\hat{\gD}_S$
            }
        }
    }
    Update policy parameter $\pi_\theta^k = \arg\min_\theta L(\pi_\theta^{k-1}, \hat{\gD}_S)$

    Policy model $\pi = \pi_\theta^k$
}
\label{algo:self_improving}
\end{algorithm*}
\section{Experiments}
\label{sec: exp}
\subsection{Benchmark Evaluation}\label{sec: benchmark eval}

\definecolor{front-color}{HTML}{F5FFFA}
\begin{table*}[t!]
\centering
\begin{tabular}{lcc|c|c}
\toprule
Model & \#Stepwise& \#Annotation & \#\texttt{SYN} & GSM8K \\
\midrule
LLaMA-2 7B & -- & 0 & 0 & 14.6 \\
+ AlphaLLM & \ding{53} & 7.5k & 8.6k & 26.5\\
+ AlphaLLM-Q & \checkmark & 7.5k & 8.6k & 31.7  \\
+ AlphaLLM-CO  & \ding{53} & 7.5k & 8.6k & 29.7  \\
+ AlphaLLM-PL-Shuffle (epoch 1) & \checkmark & 7.5k & 8.6k & 32.4  \\
+ AlphaLLM-PL-Shuffle (epoch 2) & \checkmark & 7.5k & 8.6k & 32.1  \\
\rowcolor{gray!30}
+ \ours (epoch 1) & \checkmark & 7.5k & 8.6k & 34.3  \\
\rowcolor{gray!30}
+ \ours (epoch 2) & \checkmark & 7.5k & 8.6k & 36.5  \\
\midrule
Mistral 7B & -- & 0  & 0 & 38.5 \\
+ AlphaLLM & \ding{53} & 7.5k  & 10k & 46.7  \\
+ AlphaLLM-Q & \checkmark & 7.5k  & 10k & 53.9  \\
+ AlphaLLM-CO & \ding{53} & 7.5k  & 10k & 52.3  \\
+ AlphaLLM-PL-Shuffle (epoch 1) & \checkmark & 7.5k  & 10k & 54.8  \\
+ AlphaLLM-PL-Shuffle (epoch 2) & \checkmark & 7.5k  & 10k & 53.6  \\
\rowcolor{gray!30}
+ \ours (epoch 1) & \checkmark & 7.5k  & 10k & 55.8 \\
\rowcolor{gray!30}
+ \ours (epoch 2) & \checkmark & 7.5k  & 10k & 57.3 \\
\midrule
\midrule
Model & \#Stepwise& \#Annotation & \#\texttt{SYN} & MATH \\
\midrule
Llama3 8B-instruct & -- & 0 & -- & 28.2 \\
+ AlphaLLM & \ding{53} & 7.5k &  0 &  31.0 \\
+ AlphaLLM-Q & \checkmark & 7.5k &  0 & 31.4 \\
+ AlphaLLM-CO & \ding{53} & 7.5k &  0 &  31.6 \\
+ AlphaLLM-PL-Shuffle (epoch 1) & \checkmark & 7.5k  & 0 &  32.2 \\
+ AlphaLLM-PL-Shuffle (epoch 2) & \checkmark & 7.5k  & 0 &  31.7 \\
\rowcolor{gray!30}
+ \ours (epoch 1) & \checkmark & 7.5k &  0 &  32.6 \\
\rowcolor{gray!30}
+ \ours (epoch 2) & \checkmark & 7.5k &  0 &  33.1 \\
\bottomrule
\end{tabular}
\caption{Comparison results of \ours on the GSM8K and MATH datasets, utilizing LLaMA2-7B and Mistral-7B as base models for GSM8K and LLaMA3-8B-instruct for MATH, respectively.
\#Stepwise indicates whether stepwise information provided by MCTS is utilized during the training process.
\#Annotation indicates the quantity of labeled data employed during model development, including training the value networks of MCTS and finetuning the base models.
\#\texttt{SYN} represents the number of synthetic prompts used for model training, where trajectories are generated using MCTS.}
\label{tab:main_exp}
\end{table*}

\paragraph{Datasets}
\textsc{AlphaLLM-CPL} is generally applicable to a wide spectrum of reasoning tasks.
As an early exploration, here we conduct experiments on mathematical reasoning problems where the learning signals
are clear to define i.e., , final answer is correct or wrong.
We choose to evaluate on two widely used
datasets GSM8K \cite{cobbe2021training} and MATH \cite{hendrycks2measuring}.
For GSM8K, we utilize prompts from MetaMath \cite{yumetamath} for extracting pairs based on MCTS.
The whole test set of GSM8K is used for evaluation and all training data is taken to train the value network for guiding MCTS.
For MATH, we use their whole training set to generate preference pairs for preference learning. Due to computation constraints, we utilize a subset following the
same procedure of \citet{lightman2023let} for MATH evaluation.

\paragraph{Baselines}
We select three strong open-source LLMs as baselines for our experiments: \textbf{LLaMA2-7B} and \textbf{Mistral-7B} on GSM8K, and \textbf{LLaMA3-8B-instruct} on MATH.
Following the steps of AlphaLLM, we conduct MCTS on GSM8K and MATH using the corresponding LLMs, and performed SFTwith the highest value trajectory found through MCTS, resulting in three checkpoints. The three different checkpoints obtained from SFT are referred to as \textbf{AlphaLLM}, which serve as the starting checkpoints for the following preference learning. 
In the preference learning phase, we used DPO as the objective function and compared three baselines:
(1) \textbf{AlphaLLM-CO} (Complete Only): This baseline performs DPO using only the complete trajectory pairs obtained from the MCTS search.
(2) \textbf{AlphaLLM-PL-shuffle}: This method uses both complete trajectory pairs and stepwise trajectory pairs for preference learning, but the training order is randomized.
(3) \textbf{AlphaLLM-Q}: This approach is proposed by~\citet{xie2024montecarlotreesearch} which selects steps with the highest and lowest Q values as positive and negative samples at each tree depth, forming preference pairs for DPO.
For all methods, we save a checkpoint every 200 training steps and report the performance of the best one.
All experiments are conducted on 8$\times$A6000 GPUs.

\paragraph{Experiment Results}
\textbf{(1) \ours can significantly boost reasoning ability of LLMs on math tasks.} 
As shown in Table~\ref{tab:main_exp}, compared to the three LLM baselines (LLaMA2, Mistral, and LLaMA3-instruct), our method achieves significant performance improvement. The baseline models without any preference learning generally have the lowest performance, with LLAMA2 achieving a GSM8K score of 14.6, Mistral achieving 38.5, and LLAMA3-instruct reaching 28.2 on MATH. After applying all preference learning methods, \ours achieves the highest performance, with LLAMA-2 7B reaching 36.5 on GSM8K (\textbf{150\%} improvement), Mistral 7B achieving 57.3 (\textbf{48.8\%} improvement), and LLAMA3-instruct scoring 33.1 on MATH (\textbf{17.4\%} improvement). This suggests that by organizing the learning process more effectively, \ours can significantly enhance the reasoning ability of LLMs.
\textbf{(2) Both components of \ours are crucial for enhancing the reasoning ability of LLMs.} 
Compared to the other three preference learning baselines, our method demonstrates a significant advantage across the two benchmarks. First, after incorporating stepwise trajectory pairs into the training dataset, the performance of AlphaLLM-PL-Shuffle shows a marked improvement over both AlphaLLM-CO and AlphaLLM-Q, highlighting the importance of utilizing the stepwise information provided by stepwise trajectory pairs. Moreover, after applying CPL for offline training on top of AlphaLLM-PL-Shuffle, the performance of the LLM is further enhanced (by 12.6\%, 4.5\%, and 2.8\% across two different benchmarks on three base models). The experimental comparison with the other three baselines clearly demonstrates that both components of our proposed method are crucial for the final model's performance improvement.
\textbf{(3) \ours can continue to improve model performance even after multi-epoch offline training.}
We conduct two epochs of offline training on both AlphaLLM-PL-Shuffle and \ours without changing any training data. 
It can be observed that AlphaLLM-PL-Shuffle overfits during the second epoch, resulting in a performance decline, whereas CPL, due to its ability to dynamically adjust the training order and prioritize better trajectory pairs, continues to improve performance in the second epoch. This further demonstrates the importance of the training order of trajectory pairs and the effectiveness of our proposed method.

\subsection{Impact of Stepwise Trajectory Pairs}

In the Sec~\ref{sec: benchmark eval}, we experimentally demonstrate the importance of stepwise trajectory pairs in improving LLM performance. 
In this section, we will further explore the impact of the quantity of stepwise trajectory pairs on performance enhancement and the preference learning process. 
We choose Mistral-7B as the base model and conduct experiments on GSM8K using Mistral-7B after AlphaLLM SFT. 
We set the trajectory value gap margin $\tau$ between positive and negative trajectories to 1  for constructing trajectory pairs. 
Under this setting, we obtain a total of 59K complete trajectory pairs and 47K stepwise trajectory pairs.
To avoid the influence of CPL on the experimental results, we only used DPO for preference learning. 

\begin{figure}[!htbp]
\centering
\includegraphics[width=0.49\textwidth]{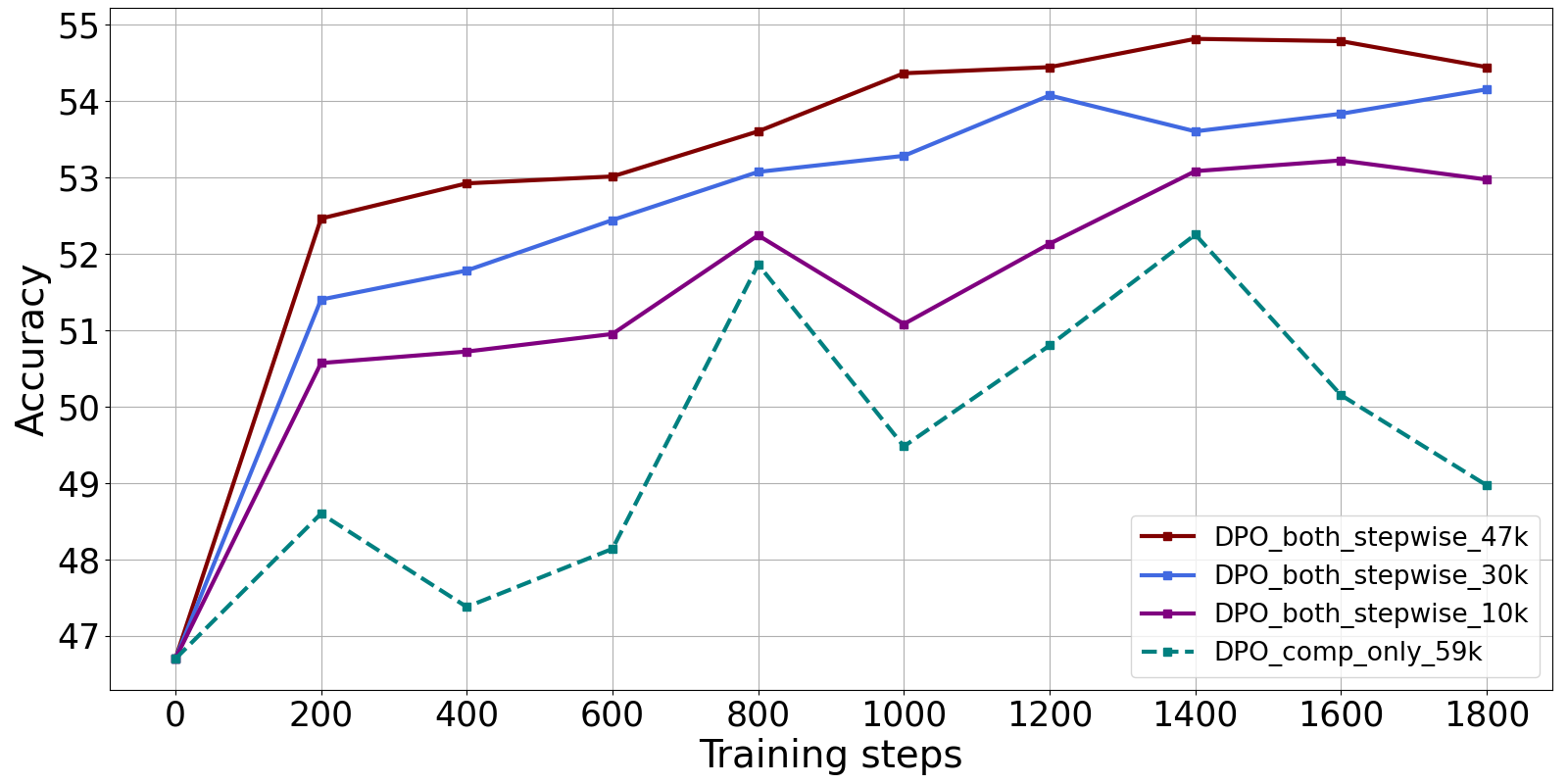}
\caption{The accuracy curve of the Mistral-7B under different stepwise trajectory pairs as the DPO training steps progress on GSM8K. The starting checkpoint is the Mistral-7B model after AlphaLLM SFT. We can observe that a larger number of stepwise trajectory pairs leads to better performance and a more stable training process.}
\label{fig:train_curve}
\end{figure}

We conduct a total of four experiments: \textbf{DPO\_comp\_only\_59k}, which uses only the complete trajectory pairs; \textbf{DPO\_both\_stepwise\_10k} and \textbf{DPO\_both\_stepwise\_30k}, where 10k and 30k stepwise trajectory pairs are randomly sampled from the 47k stepwise trajectory pairs and mixed with the complete trajectory pairs; and \textbf{DPO\_both\_stepwise\_47k}, which uses all 47k stepwise trajectory pairs mixed with the complete trajectory pairs for DPO.

In Figure~\ref{fig:train_curve}, we present the training curves showing how LLM performance changes with increasing training steps under each setting. 
From the experimental results, it is evident that when only complete trajectory pairs are used, not only is the performance improvement of the LLM insignificant, but the preference learning process is also highly unstable. 
The ACC on GSM8K fluctuates significantly, and there is a noticeable decline in performance in the later stages of training. 
In contrast, as the number of stepwise trajectory pairs increases, we observe that the preference learning process becomes progressively more stable, and the performance of the LLM improves significantly. 
This experiment further demonstrates the importance of utilizing the stepwise information from stepwise trajectory pairs.

\subsection{Ablation Study}

\begin{figure}[!htbp]
\centering
\includegraphics[width=0.49\textwidth]{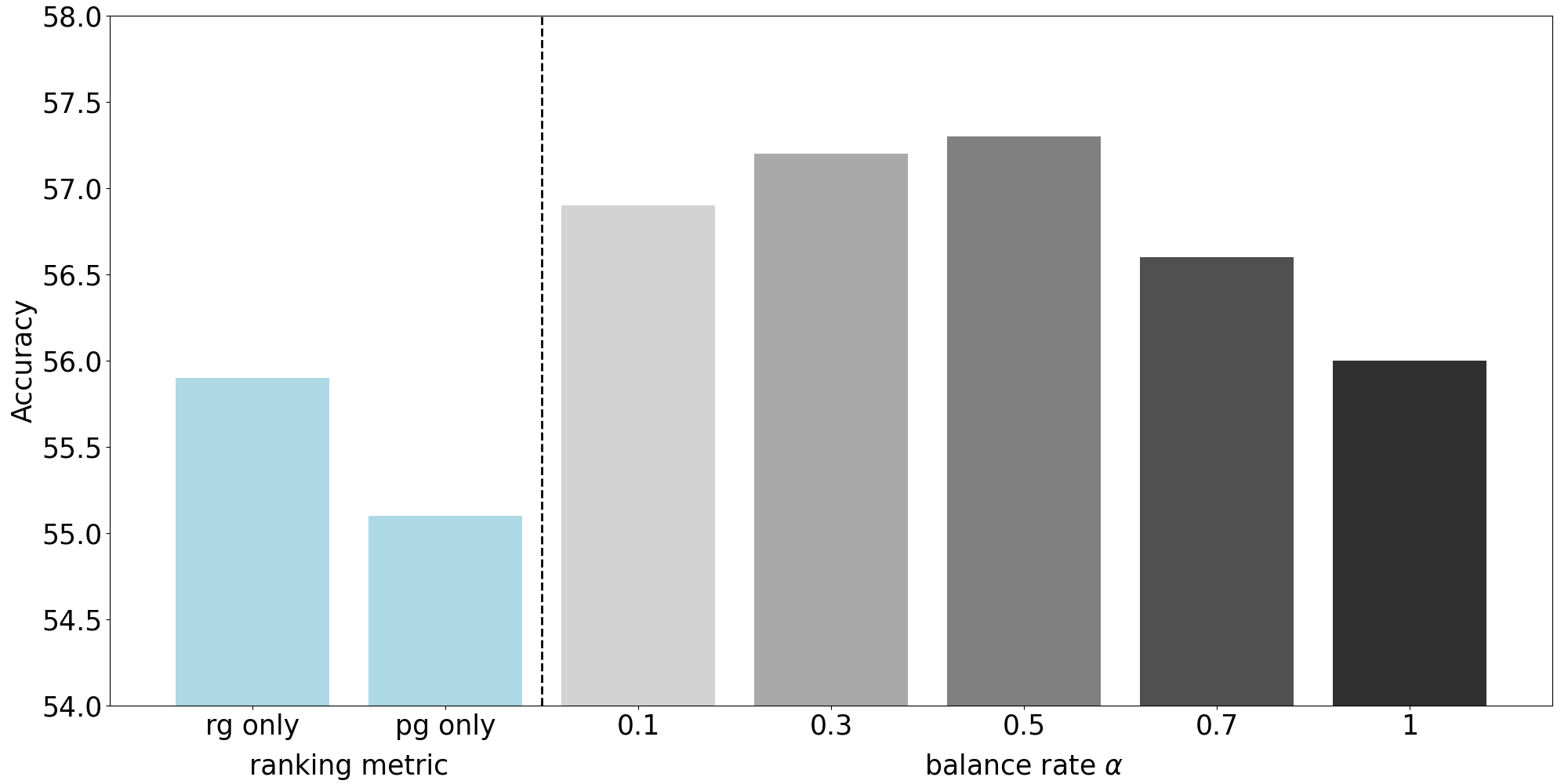}
\caption{Ablation study on balance rate and CPL metric. We can observe that the trajectory reward gap needs to play a dominant role in the metric to achieve the best performance.}
\label{fig:ablation}
\end{figure}

In this section, we conduct an ablation study to investigate the effect of the balance rate between the preference reward gap (rg) and policy prediction gap (pg) on the performance of \ours. 
We still choose Mistral-7B as the base model and conduct experiment on GSM8K.
We report the results of using only \textbf{rg} as the metric, only \textbf{pg} as the metric, as well as five different balance rates ranging from 0.1 to 1. 
All training is conducted for 2 epochs, and we report the best-performing checkpoint from epoch 2. The experimental results are shown in Figure~\ref{fig:ablation}.

Based on the results, when using only one metric, the performance with \textbf{rg} is better than with \textbf{pg}. We believe this is because the value network provides more accurate supervisory information. 
As we gradually increase the weight of \textbf{pg} in the overall metric (i.e., the balance rate), the performance of \ours also improves, reaching its peak at 0.5 with 57.3. 
After that, \ours gradually declines back to the level of using only \textbf{rg}. This experiment indicates that with a reasonable balance, \textbf{pg} can effectively improve \ours performance by providing dynamic information for adjusting trajectory pairs. 
However, since the value network obtained through LLM self-training is more accurate, \textbf{rg} should still play the dominant role in determining the training order to achieve better performance.
\section{Related works}

As the key to the success of scalable oversight \citep{bowman2022measuring},
self-improving for LLM aims to align the LLM to human preference and values mainly using the
supervision from its internal knowledge~\citep{yuan2024self,pang2024iterative,wang2024enhancing,wu2024meta}.
The most crucial part of self-improving is the critiquing signal for selecting high quality LLM responses given input queries for further model training.
Initial work \citep{bai2022constitutional,wang2023self} relies on a few hand-crafted principles or heuristic rules to filter out low-quality or redundant data.
Since it is non-trivial to reach broad coverage over different tasks, these efforts only showcase on a specific aspect, such as safety or instruction following.
Later work \citep{sun2024principle,liself,guo2024human}  composes limited number (usually hundreds) of examples, then representative in-context examples are selected to combine with general principles in order to provide more detailed background information to LLM.
They hope that the LLM can automatically designate selected in-context examples into
each data point to better guide data filtering. However, this requires the LLM to have strong abilities
to apply these principles for each specific case and make correct judgments.

Recent efforts \citep{feng2023alphazero,tian2024toward,chen2024step,xie2024monte,zhang2024rest} propose MCTS-based self-improvement, taking the outputs from MCTS to train the policy.
This is because MCTS can effectively find outputs that are a lot more accurate than vanilla outputs from the same policy model.
In terms of self-improvement methods, these efforts typically rely on either basic rejection sampling \citep{feng2023alphazero,tian2024toward,zhang2024rest} or offline preference-learning \citep{chen2024step,xie2024monte} algorithms, such as DPO \citep{rafailov2024direct}.
Different from previous
work, we propose to a more comprehensive training framework: it takes rejection sampling and curriculum preference learning to fully exploit the knowledge provided by MCTS.
We also conduct comprehensive ablation to pinpoint the essential information within MCTS for the success of self-improving, shedding some light on the future of self-improving from search.

\section{Conclusion}
We propose \ours, a novel pairwise training framework designed for the self-improvement of LLMs through MCTS behavior distillation. 
By constructing stepwise trajectory pairs from intermediate nodes in the tree search and employing our proposed curriculum preference learning to dynamically adjust the training order of trajectory pairs, we more effectively leverage the supervisory information provided by MCTS, leading to improved MCTS behavior distillation. 
Experiments on mathematical reasoning tasks demonstrate that our approach not only significantly enhances the reasoning capabilities of LLMs but also offers clear advantages over previous MCTS behavior distillation methods.

\bibliography{iclr2025_conference.bib}

\begin{thebibliography}{42}
\providecommand{\natexlab}[1]{#1}

\bibitem[{Achiam et~al.(2023)Achiam, Adler, Agarwal, Ahmad, Akkaya, Aleman,
  Almeida, Altenschmidt, Altman, Anadkat et~al.}]{achiam2023gpt}
Josh Achiam, Steven Adler, Sandhini Agarwal, Lama Ahmad, Ilge Akkaya,
  Florencia~Leoni Aleman, Diogo Almeida, Janko Altenschmidt, Sam Altman,
  Shyamal Anadkat, et~al. 2023.
\newblock Gpt-4 technical report.
\newblock \emph{arXiv preprint arXiv:2303.08774}.

\bibitem[{Auer et~al.(2002)Auer, Cesa-Bianchi, and Fischer}]{auer2002finite}
Peter Auer, Nicolo Cesa-Bianchi, and Paul Fischer. 2002.
\newblock Finite-time analysis of the multiarmed bandit problem.
\newblock \emph{Machine learning}, 47:235--256.

\bibitem[{Bai et~al.(2022)Bai, Kadavath, Kundu, Askell, Kernion, Jones, Chen,
  Goldie, Mirhoseini, McKinnon et~al.}]{bai2022constitutional}
Yuntao Bai, Saurav Kadavath, Sandipan Kundu, Amanda Askell, Jackson Kernion,
  Andy Jones, Anna Chen, Anna Goldie, Azalia Mirhoseini, Cameron McKinnon,
  et~al. 2022.
\newblock Constitutional ai: Harmlessness from ai feedback.
\newblock \emph{arXiv preprint arXiv:2212.08073}.

\bibitem[{Bowman et~al.(2022)Bowman, Hyun, Perez, Chen, Pettit, Heiner,
  Luko{\v{s}}i{\=u}t{\.e}, Askell, Jones, Chen et~al.}]{bowman2022measuring}
Samuel~R Bowman, Jeeyoon Hyun, Ethan Perez, Edwin Chen, Craig Pettit, Scott
  Heiner, Kamil{\.e} Luko{\v{s}}i{\=u}t{\.e}, Amanda Askell, Andy Jones, Anna
  Chen, et~al. 2022.
\newblock Measuring progress on scalable oversight for large language models.
\newblock \emph{arXiv preprint arXiv:2211.03540}.

\bibitem[{Chen et~al.(2024)Chen, Liao, Li, and Fan}]{chen2024step}
Guoxin Chen, Minpeng Liao, Chengxi Li, and Kai Fan. 2024.
\newblock Step-level value preference optimization for mathematical reasoning.
\newblock \emph{arXiv preprint arXiv:2406.10858}.

\bibitem[{Cobbe et~al.(2021)Cobbe, Kosaraju, Bavarian, Chen, Jun, Kaiser,
  Plappert, Tworek, Hilton, Nakano et~al.}]{cobbe2021training}
Karl Cobbe, Vineet Kosaraju, Mohammad Bavarian, Mark Chen, Heewoo Jun, Lukasz
  Kaiser, Matthias Plappert, Jerry Tworek, Jacob Hilton, Reiichiro Nakano,
  et~al. 2021.
\newblock Training verifiers to solve math word problems.
\newblock \emph{arXiv preprint arXiv:2110.14168}.

\bibitem[{Feng et~al.(2023)Feng, Wan, Wen, Wen, Zhang, and
  Wang}]{feng2023alphazero}
Xidong Feng, Ziyu Wan, Muning Wen, Ying Wen, Weinan Zhang, and Jun Wang. 2023.
\newblock Alphazero-like tree-search can guide large language model decoding
  and training.
\newblock \emph{arXiv preprint arXiv:2309.17179}.

\bibitem[{Gou et~al.(2023)Gou, Shao, Gong, Yang, Huang, Duan, Chen
  et~al.}]{gou2023tora}
Zhibin Gou, Zhihong Shao, Yeyun Gong, Yujiu Yang, Minlie Huang, Nan Duan,
  Weizhu Chen, et~al. 2023.
\newblock Tora: A tool-integrated reasoning agent for mathematical problem
  solving.
\newblock \emph{arXiv preprint arXiv:2309.17452}.

\bibitem[{Guo et~al.(2024)Guo, Yao, Shen, Wei, Zhang, Wang, and
  Liu}]{guo2024human}
Hongyi Guo, Yuanshun Yao, Wei Shen, Jiaheng Wei, Xiaoying Zhang, Zhaoran Wang,
  and Yang Liu. 2024.
\newblock Human-instruction-free llm self-alignment with limited samples.
\newblock \emph{arXiv preprint arXiv:2401.06785}.

\bibitem[{Hendrycks et~al.(2021)Hendrycks, Burns, Kadavath, Arora, Basart,
  Tang, Song, and Steinhardt}]{hendrycks2measuring}
Dan Hendrycks, Collin Burns, Saurav Kadavath, Akul Arora, Steven Basart, Eric
  Tang, Dawn Song, and Jacob Steinhardt. 2021.
\newblock Measuring mathematical problem solving with the math dataset.
\newblock \emph{Sort}, 2(4):0--6.

\bibitem[{Huang et~al.(2023)Huang, Chen, Mishra, Zheng, Yu, Song, and
  Zhou}]{huang2023large}
Jie Huang, Xinyun Chen, Swaroop Mishra, Huaixiu~Steven Zheng, Adams~Wei Yu,
  Xinying Song, and Denny Zhou. 2023.
\newblock Large language models cannot self-correct reasoning yet.
\newblock In \emph{The Twelfth International Conference on Learning
  Representations}.

\bibitem[{Katharopoulos and Fleuret(2018)}]{katharopoulos2018not}
Angelos Katharopoulos and Fran{\c{c}}ois Fleuret. 2018.
\newblock Not all samples are created equal: Deep learning with importance
  sampling.
\newblock In \emph{International conference on machine learning}, pages
  2525--2534. PMLR.

\bibitem[{Kojima et~al.(2022)Kojima, Gu, Reid, Matsuo, and
  Iwasawa}]{kojima2022large}
Takeshi Kojima, Shixiang~Shane Gu, Machel Reid, Yutaka Matsuo, and Yusuke
  Iwasawa. 2022.
\newblock Large language models are zero-shot reasoners.
\newblock \emph{Advances in neural information processing systems},
  35:22199--22213.

\bibitem[{Lee et~al.(2023)Lee, Hunter, and Ruiz}]{lee2023platypus}
Ariel~N Lee, Cole~J Hunter, and Nataniel Ruiz. 2023.
\newblock Platypus: Quick, cheap, and powerful refinement of llms.
\newblock \emph{arXiv preprint arXiv:2308.07317}.

\bibitem[{Li et~al.(2024{\natexlab{a}})Li, Wang, Hu, Wei, Zheng, Hu, Zhang, and
  Peng}]{li2024common}
Chen Li, Weiqi Wang, Jingcheng Hu, Yixuan Wei, Nanning Zheng, Han Hu, Zheng
  Zhang, and Houwen Peng. 2024{\natexlab{a}}.
\newblock Common 7b language models already possess strong math capabilities.
\newblock \emph{arXiv preprint arXiv:2403.04706}.

\bibitem[{Li et~al.(2024{\natexlab{b}})Li, Yu, Zhou, Schick, Levy, Zettlemoyer,
  Weston, and Lewis}]{liself}
Xian Li, Ping Yu, Chunting Zhou, Timo Schick, Omer Levy, Luke Zettlemoyer,
  Jason~E Weston, and Mike Lewis. 2024{\natexlab{b}}.
\newblock Self-alignment with instruction backtranslation.
\newblock In \emph{The Twelfth International Conference on Learning
  Representations}.

\bibitem[{Lightman et~al.(2023)Lightman, Kosaraju, Burda, Edwards, Baker, Lee,
  Leike, Schulman, Sutskever, and Cobbe}]{lightman2023let}
Hunter Lightman, Vineet Kosaraju, Yura Burda, Harri Edwards, Bowen Baker, Teddy
  Lee, Jan Leike, John Schulman, Ilya Sutskever, and Karl Cobbe. 2023.
\newblock Let's verify step by step.
\newblock \emph{arXiv preprint arXiv:2305.20050}.

\bibitem[{Luo et~al.(2023)Luo, Sun, Xu, Zhao, Lou, Tao, Geng, Lin, Chen, and
  Zhang}]{luo2023wizardmath}
Haipeng Luo, Qingfeng Sun, Can Xu, Pu~Zhao, Jianguang Lou, Chongyang Tao, Xiubo
  Geng, Qingwei Lin, Shifeng Chen, and Dongmei Zhang. 2023.
\newblock Wizardmath: Empowering mathematical reasoning for large language
  models via reinforced evol-instruct.
\newblock \emph{arXiv preprint arXiv:2308.09583}.

\bibitem[{Mishra et~al.(2022)Mishra, Finlayson, Lu, Tang, Welleck, Baral,
  Rajpurohit, Tafjord, Sabharwal, Clark et~al.}]{mishra2022lila}
Swaroop Mishra, Matthew Finlayson, Pan Lu, Leonard Tang, Sean Welleck, Chitta
  Baral, Tanmay Rajpurohit, Oyvind Tafjord, Ashish Sabharwal, Peter Clark,
  et~al. 2022.
\newblock Lila: A unified benchmark for mathematical reasoning.
\newblock In \emph{Proceedings of the 2022 Conference on Empirical Methods in
  Natural Language Processing}, pages 5807--5832.

\bibitem[{Pang et~al.(2024)Pang, Yuan, Cho, He, Sukhbaatar, and
  Weston}]{pang2024iterative}
Richard~Yuanzhe Pang, Weizhe Yuan, Kyunghyun Cho, He~He, Sainbayar Sukhbaatar,
  and Jason Weston. 2024.
\newblock Iterative reasoning preference optimization.
\newblock \emph{arXiv preprint arXiv:2404.19733}.

\bibitem[{Rafailov et~al.(2024)Rafailov, Sharma, Mitchell, Manning, Ermon, and
  Finn}]{rafailov2024direct}
Rafael Rafailov, Archit Sharma, Eric Mitchell, Christopher~D Manning, Stefano
  Ermon, and Chelsea Finn. 2024.
\newblock Direct preference optimization: Your language model is secretly a
  reward model.
\newblock \emph{Advances in Neural Information Processing Systems}, 36.

\bibitem[{Schaul(2015)}]{schaul2015prioritized}
Tom Schaul. 2015.
\newblock Prioritized experience replay.
\newblock \emph{arXiv preprint arXiv:1511.05952}.

\bibitem[{Schulman et~al.(2017)Schulman, Wolski, Dhariwal, Radford, and
  Klimov}]{schulman2017proximal}
John Schulman, Filip Wolski, Prafulla Dhariwal, Alec Radford, and Oleg Klimov.
  2017.
\newblock Proximal policy optimization algorithms.
\newblock \emph{arXiv preprint arXiv:1707.06347}.

\bibitem[{Sun et~al.(2024)Sun, Shen, Zhou, Zhang, Chen, Cox, Yang, and
  Gan}]{sun2024principle}
Zhiqing Sun, Yikang Shen, Qinhong Zhou, Hongxin Zhang, Zhenfang Chen, David
  Cox, Yiming Yang, and Chuang Gan. 2024.
\newblock Principle-driven self-alignment of language models from scratch with
  minimal human supervision.
\newblock \emph{Advances in Neural Information Processing Systems}, 36.

\bibitem[{Team et~al.(2023)Team, Anil, Borgeaud, Wu, Alayrac, Yu, Soricut,
  Schalkwyk, Dai, Hauth et~al.}]{team2023gemini}
Gemini Team, Rohan Anil, Sebastian Borgeaud, Yonghui Wu, Jean-Baptiste Alayrac,
  Jiahui Yu, Radu Soricut, Johan Schalkwyk, Andrew~M Dai, Anja Hauth, et~al.
  2023.
\newblock Gemini: a family of highly capable multimodal models.
\newblock \emph{arXiv preprint arXiv:2312.11805}.

\bibitem[{Tian et~al.(2024)Tian, Peng, Song, Jin, Yu, Mi, and
  Yu}]{tian2024toward}
Ye~Tian, Baolin Peng, Linfeng Song, Lifeng Jin, Dian Yu, Haitao Mi, and Dong
  Yu. 2024.
\newblock Toward self-improvement of llms via imagination, searching, and
  criticizing.
\newblock \emph{arXiv preprint arXiv:2404.12253}.

\bibitem[{Tong et~al.(2024)Tong, Zhang, Wang, Wu, and
  He}]{tong2024dartmathdifficultyawarerejectiontuning}
Yuxuan Tong, Xiwen Zhang, Rui Wang, Ruidong Wu, and Junxian He. 2024.
\newblock \href {https://arxiv.org/abs/2407.13690} {Dart-math: Difficulty-aware
  rejection tuning for mathematical problem-solving}.
\newblock \emph{Preprint}, arXiv:2407.13690.

\bibitem[{Touvron et~al.(2023)Touvron, Martin, Stone, Albert, Almahairi,
  Babaei, Bashlykov, Batra, Bhargava, Bhosale et~al.}]{touvron2023llama}
Hugo Touvron, Louis Martin, Kevin Stone, Peter Albert, Amjad Almahairi, Yasmine
  Babaei, Nikolay Bashlykov, Soumya Batra, Prajjwal Bhargava, Shruti Bhosale,
  et~al. 2023.
\newblock Llama 2: Open foundation and fine-tuned chat models.
\newblock \emph{arXiv preprint arXiv:2307.09288}.

\bibitem[{Valmeekam et~al.(2024)Valmeekam, Marquez, Olmo, Sreedharan, and
  Kambhampati}]{valmeekam2024planbench}
Karthik Valmeekam, Matthew Marquez, Alberto Olmo, Sarath Sreedharan, and
  Subbarao Kambhampati. 2024.
\newblock Planbench: An extensible benchmark for evaluating large language
  models on planning and reasoning about change.
\newblock \emph{Advances in Neural Information Processing Systems}, 36.

\bibitem[{Wang et~al.(2024{\natexlab{a}})Wang, Song, Tian, Peng, Yu, Mi, Su,
  and Yu}]{wang2024litesearch}
Ante Wang, Linfeng Song, Ye~Tian, Baolin Peng, Dian Yu, Haitao Mi, Jinsong Su,
  and Dong Yu. 2024{\natexlab{a}}.
\newblock Litesearch: Efficacious tree search for llm.
\newblock \emph{arXiv preprint arXiv:2407.00320}.

\bibitem[{Wang et~al.(2024{\natexlab{b}})Wang, Chen, Wang, Zhou, Zhou, Yao,
  Zhou, Goldstein, Bhatia, Huang et~al.}]{wang2024enhancing}
Xiyao Wang, Jiuhai Chen, Zhaoyang Wang, Yuhang Zhou, Yiyang Zhou, Huaxiu Yao,
  Tianyi Zhou, Tom Goldstein, Parminder Bhatia, Furong Huang, et~al.
  2024{\natexlab{b}}.
\newblock Enhancing visual-language modality alignment in large vision language
  models via self-improvement.
\newblock \emph{arXiv preprint arXiv:2405.15973}.

\bibitem[{Wang et~al.(2023{\natexlab{a}})Wang, Wongkamjan, Jia, and
  Huang}]{wang2023live}
Xiyao Wang, Wichayaporn Wongkamjan, Ruonan Jia, and Furong Huang.
  2023{\natexlab{a}}.
\newblock Live in the moment: Learning dynamics model adapted to evolving
  policy.
\newblock In \emph{International Conference on Machine Learning}, pages
  36470--36493. PMLR.

\bibitem[{Wang et~al.(2023{\natexlab{b}})Wang, Kordi, Mishra, Liu, Smith,
  Khashabi, and Hajishirzi}]{wang2023self}
Yizhong Wang, Yeganeh Kordi, Swaroop Mishra, Alisa Liu, Noah~A Smith, Daniel
  Khashabi, and Hannaneh Hajishirzi. 2023{\natexlab{b}}.
\newblock Self-instruct: Aligning language models with self-generated
  instructions.
\newblock In \emph{The 61st Annual Meeting Of The Association For Computational
  Linguistics}.

\bibitem[{Wei et~al.(2022)Wei, Wang, Schuurmans, Bosma, Xia, Chi, Le, Zhou
  et~al.}]{wei2022chain}
Jason Wei, Xuezhi Wang, Dale Schuurmans, Maarten Bosma, Fei Xia, Ed~Chi, Quoc~V
  Le, Denny Zhou, et~al. 2022.
\newblock Chain-of-thought prompting elicits reasoning in large language
  models.
\newblock \emph{Advances in neural information processing systems},
  35:24824--24837.

\bibitem[{Wu et~al.(2024)Wu, Yuan, Golovneva, Xu, Tian, Jiao, Weston, and
  Sukhbaatar}]{wu2024meta}
Tianhao Wu, Weizhe Yuan, Olga Golovneva, Jing Xu, Yuandong Tian, Jiantao Jiao,
  Jason Weston, and Sainbayar Sukhbaatar. 2024.
\newblock Meta-rewarding language models: Self-improving alignment with
  llm-as-a-meta-judge.
\newblock \emph{arXiv preprint arXiv:2407.19594}.

\bibitem[{Xie et~al.(2024{\natexlab{a}})Xie, Goyal, Zheng, Kan, Lillicrap,
  Kawaguchi, and Shieh}]{xie2024montecarlotreesearch}
Yuxi Xie, Anirudh Goyal, Wenyue Zheng, Min-Yen Kan, Timothy~P. Lillicrap, Kenji
  Kawaguchi, and Michael Shieh. 2024{\natexlab{a}}.
\newblock \href {https://arxiv.org/abs/2405.00451} {Monte carlo tree search
  boosts reasoning via iterative preference learning}.
\newblock \emph{Preprint}, arXiv:2405.00451.

\bibitem[{Xie et~al.(2024{\natexlab{b}})Xie, Goyal, Zheng, Kan, Lillicrap,
  Kawaguchi, and Shieh}]{xie2024monte}
Yuxi Xie, Anirudh Goyal, Wenyue Zheng, Min-Yen Kan, Timothy~P Lillicrap, Kenji
  Kawaguchi, and Michael Shieh. 2024{\natexlab{b}}.
\newblock Monte carlo tree search boosts reasoning via iterative preference
  learning.
\newblock \emph{arXiv preprint arXiv:2405.00451}.

\bibitem[{Yu et~al.(2023)Yu, Jiang, Shi, Jincheng, Liu, Zhang, Kwok, Li,
  Weller, and Liu}]{yumetamath}
Longhui Yu, Weisen Jiang, Han Shi, YU~Jincheng, Zhengying Liu, Yu~Zhang, James
  Kwok, Zhenguo Li, Adrian Weller, and Weiyang Liu. 2023.
\newblock Metamath: Bootstrap your own mathematical questions for large
  language models.
\newblock In \emph{The Twelfth International Conference on Learning
  Representations}.

\bibitem[{Yuan et~al.(2024)Yuan, Pang, Cho, Sukhbaatar, Xu, and
  Weston}]{yuan2024self}
Weizhe Yuan, Richard~Yuanzhe Pang, Kyunghyun Cho, Sainbayar Sukhbaatar, Jing
  Xu, and Jason Weston. 2024.
\newblock Self-rewarding language models.
\newblock \emph{arXiv preprint arXiv:2401.10020}.

\bibitem[{Yuan et~al.(2023)Yuan, Yuan, Li, Dong, Tan, and
  Zhou}]{yuan2023scaling}
Zheng Yuan, Hongyi Yuan, Chengpeng Li, Guanting Dong, Chuanqi Tan, and Chang
  Zhou. 2023.
\newblock Scaling relationship on learning mathematical reasoning with large
  language models.
\newblock \emph{arXiv preprint arXiv:2308.01825}.

\bibitem[{Yue et~al.(2023)Yue, Qu, Zhang, Fu, Huang, Sun, Su, and
  Chen}]{yue2023mammoth}
Xiang Yue, Xingwei Qu, Ge~Zhang, Yao Fu, Wenhao Huang, Huan Sun, Yu~Su, and
  Wenhu Chen. 2023.
\newblock Mammoth: Building math generalist models through hybrid instruction
  tuning.
\newblock \emph{arXiv preprint arXiv:2309.05653}.

\bibitem[{Zhang et~al.(2024)Zhang, Zhoubian, Yue, Dong, and
  Tang}]{zhang2024rest}
Dan Zhang, Sining Zhoubian, Yisong Yue, Yuxiao Dong, and Jie Tang. 2024.
\newblock Rest-mcts*: Llm self-training via process reward guided tree search.
\newblock \emph{arXiv preprint arXiv:2406.03816}.

\end{thebibliography}

\appendix
\section{Hyperparameters}

In this section, we present hyperparameters used in our experiments. For trajectory pair extraction, both complete and stepwise trajectory pairs are selected for training with the margin $\tau$ as 1. In the curriculum preference learning (CPL), we set the balance rate between the trajectory reward gap and policy prediction gap to 0.5. The batch size for SFT is 128 and the batch size for preference learning is set to 64. The learning rates used at different stages are shown in Table~\ref{tab:lr}.

\begin{table*}[t!]
\centering
\begin{tabular}{lccc}
\toprule
Model & SFT& CPL-epoch 1 & CPL-epoch 2 \\
\midrule
LLaMA-2 7B & 1e-5 & 5e-7 & 5e-8 \\
Mistral 7B & 1e-7 & 2e-7 & 1e-8 \\
LLaMA-3 8B-instruct & 1e-6 & 5e-7 & 5e-8 \\
\bottomrule
\end{tabular}
\caption{Learning rate used in different stages of different LLMs.}
\label{tab:lr}
\end{table*}

\end{document}